\documentclass[10pt, a4paper]{article}
\usepackage{lrec}
\usepackage{graphicx}
\usepackage{tabularx}
\usepackage{color,soul}
\usepackage{array}
\usepackage{booktabs}
\usepackage{bbold}
\usepackage{subfigure}
\usepackage{bm}
\usepackage{url}

\usepackage{epstopdf}
\usepackage{CJKutf8}

\usepackage{hyperref}
\usepackage{xstring}

\title{A Chinese Corpus for Fine-grained Entity Typing}

\name{Chin Lee\textsuperscript{1}, Hongliang Dai\textsuperscript{1}, Yangqiu Song\textsuperscript{1}, Xin Li\textsuperscript{2}}

\address{\textsuperscript{1}HKUST
\\
\textsuperscript{2}Tencent Technology (SZ) Co., Ltd. \\
         cleeag@ust.hk, \{hdai,yqsong\}@cse.ust, alonsoli@tencent.com\\}

\abstract{
Fine-grained entity typing is a challenging task with wide applications. However, most existing datasets for this task are in English. In this paper, we introduce a corpus for Chinese fine-grained entity typing that contains 4,800 mentions manually labeled through crowdsourcing. Each mention is annotated with free-form entity types. To make our dataset useful in more possible scenarios, we also categorize all the fine-grained types into 10 general types. Finally, we conduct experiments with some neural models whose structures are typical in fine-grained entity typing and show how well they perform on our dataset. We also show the possibility of improving Chinese fine-grained entity typing through cross-lingual transfer learning. 
\\ \newline  }

\begin{document}

\maketitleabstract

\section{Introduction}
The task of fine-grained entity typing \cite{ling2012fine,gillick2014context} assigns fine-grained types such as \textit{/person/politician}, \textit{/organization/company} to entity mentions in texts.
It provides additional details to entity mentions compared with the typing in traditional named entity recognition tasks \cite{chinchor1998overview,finkel2005incorporating}, which typically categorize entity mentions into very general types such as \textit{person}, \textit{location}, or \textit{organization}.

Ultra-fine Entity Typing \cite{choi2018ultra} introduces a new fine-grained entity typing task that requires to predict an open set of types for entity mentions. The dataset constructed for this task uses a very large tag set that contains around 10k free-form type phrases, while previous fine-grained entity typing datasets usually use tag sets with no greater than 200 types. This task presents a much closer view for each entity mention. Consider the sentence: ``Tim Cook announced the new iPhone this morning.'' With the dataset constructed by \cite{gillick2014context}, the mention ``Tim Cook'' can only be identified as \textit{/person/business}. 
But with ultra-fine entity typing,
``Tim Cook'' can be categorized under types such as \textit{businessman}, \textit{executive}, \textit{public figure}, etc.
These free-form type phrases provide a more comprehensive and detailed description on the entity mention.


\begin{CJK*}{UTF8}{gbsn}
\begin{table}[t!]
		\begin{center}
        {\small
			\begin{tabular}{p{4.2cm}|p{3.1cm}}
				\toprule Sentence with Mention & Label Types \\ \midrule
				高尔基大街（现易名为\textbf{\textcolor{blue}{特维尔大街}}）是莫斯科一条最主要的大街
				Gorky Street (now as known as \textbf{\textcolor{blue}{Tverskaya Street}}) is one of the main streets in Moscow. & 街道/street, 路/road, 旅游景点/tourist attraction, 街/street, 大街/thoroughfare, 道路/ path\\ 
				\hline
				\textbf{\textcolor{blue}{腾讯}}、天猫或许将成为最大的受益者。 
				
				\textbf{\textcolor{blue}{Tencent}}, TMall may benefit the most.&品牌/brand, 公司/ company \\
				\hline
				
				\textbf{\textcolor{blue}{欧佩克}}去年11月份决定今年上半年该组织原油日产限额从2503万桶提高到2750万桶。 
				
				\textbf{\textcolor{blue}{OPEC}} decided to increase the limit of daily production unit for the organization.& 国际组织/ international organization, 组织/organization, 联盟/league \\
				\hline
				我在\textbf{\textcolor{blue}{西堤}}牛排上海虹口龙之梦店：同学小聚∩∩哈哈
				
				I'm at \textbf{\textcolor{blue}{Tasty}} Shanghai store: Friends gathering, haha & 品牌/ brand, 地方/ location, 餐馆/ restaurant, 位置/ location\\
				
				\hline
				嘿嘿，比赛前厚着脸皮拉着\textbf{\textcolor{blue}{顾老师}}合了好几张嘿嘿
				
				haha, took some pictures with \textbf{\textcolor{blue}{Mr. Gu}} before the game, haha & 人/person, 老师/teacher, 教师/ school teacher\\

				\bottomrule
			\end{tabular}
        }
		\end{center}
\caption{\label{tab-crowd-samp}Samples from our crowdsourced dataset. Each example contains an entity mention, the context sentence, and the annotated labels. The entity mentions are highlighted in blue. The first three rows are from news or magazines; the last two rows are from Weibo, a Chinese social media platform similar to Twitter.}
\end{table}
\end{CJK*}

Unfortunately, most corpora \cite{ling2012fine,weischedel2005,gillick2014context,choi2018ultra} of fine-grained entity typing are in English. To our knowledge, there doesn't exist a large-scale fine-grained entity typing dataset exclusively in Chinese. In view of the growth of the research in Chinese NLP, a dataset for Chinese fine-grained entity typing will provide great value. Thus, in this paper, we present a Chinese corpus of extremely fine-grained entity typing containing over 7,100 unique entity types. We adopt a similar policy as the Ultra-fine Entity Typing corpus \cite{choi2018ultra} by allowing an open set of entity types for each entity mention. In addition, we construct 10 general types, and mapped each fine-grained type to them. This provides a simple hierarchy and can also be useful for downstream tasks.

Our dataset consists of two parts: a relatively small set of examples annotated via crowdsourcing that contains 4,800 entity mention examples, and a large corpus annotated via distant supervision with 1.9M entity mentions. The former is accurate and can be used for both training and evaluation; the latter can only be used for training. Different from the dataset in \cite{choi2018ultra}, in addition to news, magazines and web articles, we also include samples from social media which contains informal texts. Table \ref{tab-crowd-samp} lists some examples from our crowdsourced dataset.

Our code and dataset are available at \url{https://github.com/HKUST-KnowComp/cfet}.

\section{Dataset Construction}

We annotate an open type set for each entity mention with a procedure similar to the Ultra-Fine Entity Typing task \cite{choi2018ultra}. This annotation procedure benefits from having greater overall type coverage, and the types also produce a more comprehensive description for each of the entity mentions.
Our dataset is generated with two different methods: crowdsourcing via Amazon Mechanical Turk, and entity linking between Wikipedia and Wikidata. Crowdsourcing can provide an accurate dataset for both training and evaluation, distant supervision via entity linking can create a large corpus for training. On top of that, we provide a mapping between the fine-grained types and the 10 general types defined by us.


\subsection{Annotation Via Crowdsourcing}

We gather our entity mentions from four different sources: Golden Horse \cite{HeS16}, Boson dataset provided by BosonNLP\footnote{\url{https://bosonnlp.com/dev/resource}}, MSRA's open source NER dataset\footnote{\url{https://www.microsoft.com/en-us/download/details.aspx?id=52531}}, and PKU's Corpus of Multi-level Processing for Modern Chinese \cite{DVN/SEYRX5_2018}. Each source has distinct semantic and lexical characteristics, which ensures the diversity of the dataset. For the Boson, MSRA and PKU's dataset, the sentences are mostly extracted from news or magazines, and thus are more formal and detailed. For the Golden Horse dataset, most of them are extracted from Weibo (a Chinese social media website similar to Twitter) posts, which are far more informal. We extract mentions from these sources and amass around 4,800 entity mentions with context sentences. 80\% of the mentions are named entities 
\begin{CJK*}{UTF8}{gbsn}
(e.g.香港/\textit{Hong Kong}, 苹果公司/\textit{Apple Inc.}, 勒布朗-詹姆斯/\textit{LeBron James}) and 20\% of them are pronouns.
\end{CJK*}

Our crowdsourcing procedure consists of two steps. In step one, we let the annotators annotate entity mentions based on a type vocabulary we provide. The type vocabulary is constructed with types extracted from Wikidata and types provided by Ultra-Fine Entity Typing \cite{choi2018ultra}. It contains around 14K distinct types. We also provide a mapping from simplified Chinese to both English and Traditional Chinese and let the annotators decide which language to use. We require 3 different annotators to annotate 2 types for each entity mention, i.e., there will be at most 6 distinct labels for each entity mention. Similar to previous work \cite{gillick2014context}, the label for each entity mention should be context dependent. If an entity mention has many eligible types (e.g., Donald Trump can be politician, businessmen, or television host), we ask the annotators to annotate the types that most closely reflect the context. If the context does not provide any relevant information for annotating the mention, the annotators are asked to label them with the most well-known types at their discretion.

\begin{figure}[t]
\includegraphics[width=.5\textwidth]{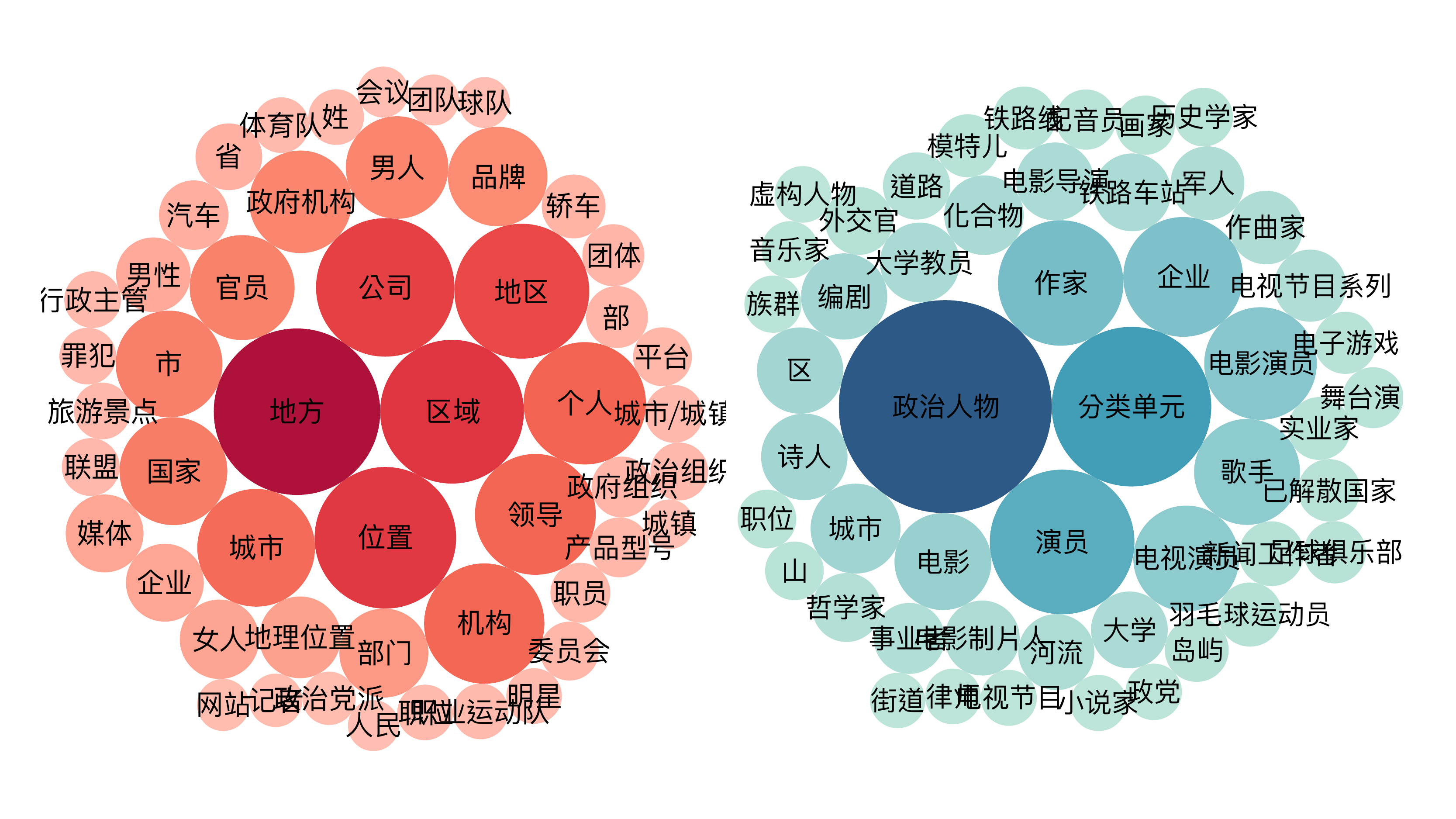}
\caption{\label{bubble-f}Visualization of the top 50 occurring fine-grained types. Left: crowdsourced dataset. Right: distant supervision dataset. The area of each bubble corresponds to its occurrence in the dataset.}
\end{figure}

In step two, we present all the types annotated for each entity mention in step one and let five different annotators determine if each type of annotation is valid or not. We analyze this validation result and find that each pair of annotators agreed on 67.2\% of the validation results they made. The disagreements result from a different understanding of certain entity terms, on the task definition, and on whether an entity belongs to a type. Our final dataset consists of only the types approved by more than 3/5 of the annotators. In total, we obtain around 4,800 unique examples and 1,300 unique types. The left side of Figure \ref{bubble-f} shows the 50 most occurring fine-grained types in this dataset.

\subsection{Annotation Via Distant Supervision}

We construct our distant supervision dataset with the combination of Wikipedia and Wikidata. Inspired by prior work \cite{ling2012fine,mintz-etal-2009-distant}, we use the anchor links in the Wikipedia data as our entity mentions. We explore all the items (each item in Wikidata may corresponds to an entity) in Wikidata and select those with a Chinese Wikipedia page as possible entities. Since each Wikipedia page title is unique, we can then link the entity mentions from Wikipedia to Wikidata and utilize the fields and properties in Wikidata to obtain the types for each mention. For each entity in Wikidata, we take the following properties as their types: instance of, subclass of, and occupation. For example, Leonardo DiCaprio has an instance of \textit{human}, with occupations of \textit{actor, film actor, screenwriter, television actor, film producer,} and \textit{stage actor}. This distantly annotates an entity mention with types, and we can extract its context sentence to form a training sample. In total, we gather 1.9M training examples and 5,975 unique types with this approach. The 50 most occurring fine-grained types in this dataset is shown in the right side of Figure \ref{bubble-f}. Although a large number of samples can be obtained this way, it has the limitation that the labeled types for an entity mention do not reflect the context. Also, each entity mention normally possesses less then 3 fine-grained types.

      

\begin{figure}[h!]
\includegraphics[width=.45\textwidth]{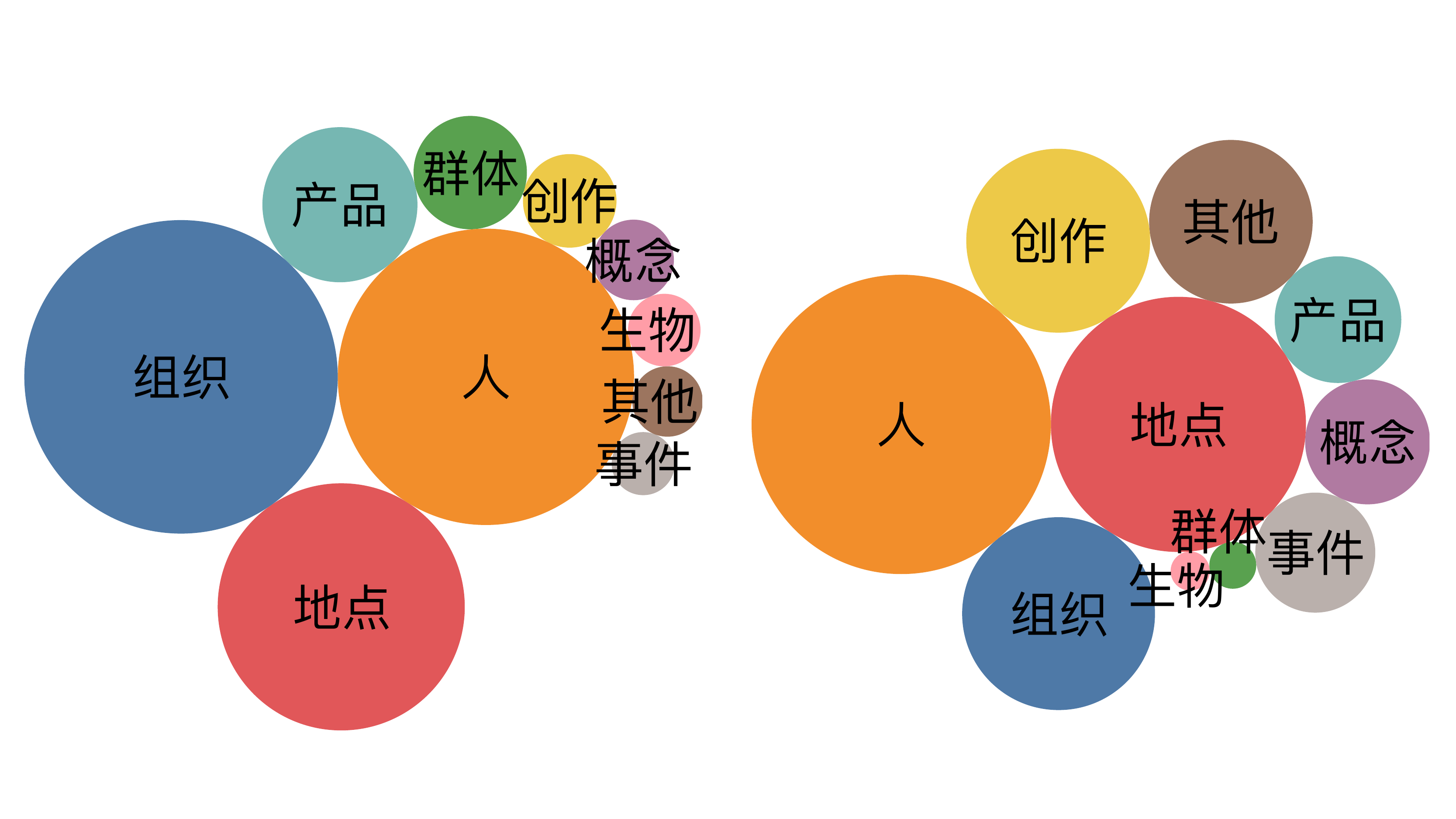}
\caption{\label{bubble-g}Bubble chart of general types. Left: crowdsourced dataset. Right: distant supervision dataset.The area of each bubble corresponds to its occurrence in the dataset.}
\end{figure}

\subsection{General Type Mapping}

Both our crowdsourced and distant supervision method provide great varieties of fine-grained types. However, we also believe that assigning a high-level, more general type to each entity mention is a necessity, since it may be required by some certain applications. 
Thus, all the fine-grained types are categorized into 10 general types defined by us: 
\begin{CJK*}{UTF8}{gbsn}
人/\textit{person}, 生物/\textit{living thing}, 组织/\textit{organization}, 地点/\textit{location}, 创作/\textit{creation}, 事件/\textit{event}, 概念/\textit{concept}, 产品/\textit{goods}, 群体/\textit{group}, and 其他/\textit{others}.
\end{CJK*}

In order to find the corresponding general type for each fine-grained type, we first use the type hierarchy provided in Wikidata to perform automatic type mapping. A large number of the fine-grained types in our dataset are from Wikidata, where we can find properties such as \textit{subclass of} and \textit{instance of} for them. The values of these properties are usually higher-level types. For example, the type ``company'' is a \textit{subclass of} ``organization''. Thus, we first manually assign a number of relatively coarse-grained types in Wikidata to our 10 general types. Then, for each fine-grained type in our dataset that can be found in Wikidata, we recursively search through its higher-level types to find a general type for it. This approach also introduces noise, so some mappings may be incorrect.

Finally, we manually inspected all the type mappings and fix the incorrect ones to ensure quality. Out of 7182 mappings, we found 1516 incorrect ones. Table \ref{tab-fgtc} shows the number of fine-grained types in each general type. On average, in our crowdsourced dataset, each mention has 3.1 fine-grained entity types and 1.3 general types. In our distant supervision dataset, each mention has around 1.6 fine-grained types, and 1 general type. Figure \ref{bubble-g} shows the visualization of the occurrence of general types in our datasets.

\begin{CJK*}{UTF8}{gbsn}
\begin{table}[h!]
		\begin{center}
        {
			\begin{tabular}{p{2cm}|p{1cm}|p{4cm}}
				\toprule GT & \#FGT & FGT Examples\\ \midrule

person & 1305& 交易员/trader, 女儿/ daughter, 地质学家/ geologist\\ 
\hline
living thing &98& 梨/pear,狗/dog, 象/ elephant\\
\hline
location &917&  住宅/residence, 地区首府/district capital, 胡同/ hutong \\
\hline
organization &651& 中学/secondary school, 银行/bank， 医院/hospital\\
\hline
group &45& 群众/community, 原住民/indigenous people\\
\hline
event &686& 意外事故/accident, 经济危机/economic crisis\\
\hline
concept &735& 时间/time, 经济理论/theoretical economics\\
\hline
creation &824& 社论/editorial, 文件/file, 世界地图/world map风俗艺术/genre art\\
\hline
goods &1273& 打字机/typewriter, 菜肴/dish, 电脑/computer\\
\hline
others &648& 青霉素/ penicillin, 非蛋白胺基酸/ non-proteinogenic amino acids\\
				
				\bottomrule
			\end{tabular}
        }
		\end{center}
\caption{\label{tab-fgtc}Number and examples of fine-grained types in each general type. ``GT'' denotes general type; ``FGT'' denotes fine-grained type.}
\end{table}
\end{CJK*}

\begin{table}[h!]
		\begin{center}
			\begin{tabular}{cccc}
				\toprule 
				Dataset & Crowdsourced & Distant \\ 
				\midrule
				Mentions & 4,798 & 1,908,481\\
				Unique FGT & 1,307 & 5,975\\
			    GT per mention & 1.6 & 1.0\\
				FGT per mention & 3.1 & 1.3\\
				\bottomrule
			\end{tabular}
		\end{center}
\caption{\label{tab:stat} Statistics for our crowdsourced and distant dataset.}
\end{table}

\begin{table*}[t!]
\begin{center}
\begin{tabular}{c|c|c|c|c|c|c|c|c}
\toprule
Dataset & \multicolumn{4}{c|}{Our dataset} & \multicolumn{4}{c}{Ultra-fine dataset} \\ \hline
Method & MRR & P & R & F1 & MRR & P & R & F1\\ \midrule

BiLSTM & 0.199 & 30.5 & 14.6 & 19.8 & 0.160 & 27.0 & 16.2 & 20.3 \\
BiLSTM + General Types & 0.200 & 46.6 & 17.5 & 25.5 & - & - & - & -\\

\hline
 BERT & 0.281 & 42.2 & 30.9 & 35.7 & 0.221 & 47.9 & 20.6 & 28.8 \\
 BERT + General Types & 0.310 & 64.1 & 38.2 & 47.9 & - & - & - & - \\
\bottomrule
\end{tabular}
\end{center}
\caption{\label{tab:fet-perf} Fine-grained entity typing performance on the test set. We report mean reciprocal rank (MRR), macro-averaged precision, recall and F1 score. ``+ General Types" indicates adding the general type mapping.
}
\end{table*}

\section{Experiments}

Experiments are conducted with neural entity typing models that follow the design of previous works \cite{dai-etal-2019-improving,shimaoka-etal-2016-attentive}. We experimented with structures such as bi-LSTM and BERT \cite{devlin2019bert}. We also trained both models on the Ultra fine-grained dataset \cite{choi2018ultra} for comparison.

\subsection{Experimental Settings}

Similar to the typical neural entity typing models, the architecture of the models we experimented consist of three parts: context sequence representation, mention representation, and the final inference layer. We adapted certain model architectures to better match our Fine-grained typing objective. We use fastText \cite{mikolov2018advances} for Chinese word embedding and Glove 
\cite{pennington2014glove} for English word Embedding.

Both BERT implementation from HuggingFace\footnote{https://github.com/huggingface/transformers} and bidirectional LSTM are experimented to construct the context representation. Given a sentence \(x_1, ...,x_n\), we aim to construct a representation of the mention \(x_m\) with the information provided by the context in the sentence. We substitute the mention \(x_m\) with a [MASK] token and feed the whole sentence into the models. For the BiLSTMs models, we use two layers of BiLSTMs, producing output vectors \(\bm{h}^1, \bm{h}^2\). We then extract the vectors at position \(m\) from each hidden layer, and take the addition \(\bm{f}_c = \bm{h}_m^1 + \bm{h}_m^2\) as the context representation of the mention \(x_m\). Similarly, when using BERT for the context representation, we take the vector at position \(m\) in final output layer as the context sequence representation.

To construct the mention representation, we simply take average \(\bm{f}_s = (\sum_{i=1}^{l}\bm{w}_i)/l\) of the word embedding for the words in the entity mention string. We then use the concatenation \([\bm{f}_c;\bm{f}_s]\) as our input to a dense layer and obtain the output.

Following previous work \cite{dai-etal-2019-improving,yogatama2015embedding}, we assign each type a vector and compute its dot product with the output of dense layer as the score for each type. A type is predicted if its score is greater than 0. If none of the types is, we pick the type with the greatest global score.

Also similar to previous works \cite{dai-etal-2019-improving,abhishek-etal-2017-fine}, we use a customized hinge loss that better reflects the training objective of our data. When training with the general types on our dataset, or training on the Ultra-fine dataset which contain different level of granularity of types, we use a multitask objective function:

\begin{equation}
J = \sum_i J_i \cdot \mathbb{1}_i(t).
\end{equation}
Here \(i\) indicates the level of granularity. For the Ultra-fine dataset, \(i\) can be general, fine, and ultra-fine. In our dataset, \(i\) can be general or fine-grained. The input \(t\) indicates the ground truth type of a mention \(m\). We only update loss for the \(i\)th level when the ground truth contain at least one label of such level in it. Function \(J_i\) is defined as follows:

\begin{equation}
J_i = \sum_m [\sum_{t\in \tau_i}\max(0,1-s(m,t))],
\end{equation}
where  \(\tau_i\) indicates the type set for each granularity level. 
 
 \begin{CJK*}{UTF8}{gbsn}
\begin{table*}[ht]
\begin{center}
\begin{tabularx}{\textwidth}{p{0.4cm}|p{6.4cm}|p{4.3cm}|p{4.3cm}}
\toprule
      No. & Sentence & Label & Prediction\\
      \hline
      1. & \textbf{\textcolor{blue}{澳大利亚队}}夺得女子4×100米自由泳接力前三名。
      
      The \textbf{\textcolor{blue}{Australian team}} won the top three prizes for 400m freestyle women swimming.& 职业运动队/professional sports team, 团队/team, 体育队/sports team, 组织/organization, 国家队/national sports team & 职业运动队/professional sports team,团队/team, 体育队/sports,组织/organization, 国家队/national sports team\\
      
      \midrule
      \hline
      2. & \textbf{\textcolor{blue}{北京大学}}20多个院系的1000多名大学生，参加升旗仪式。
      
      More than 1000 \textbf{\textcolor{blue}{Peking University's}} students from more than 20 faculties attended the flag raising ceremony. & 教学机构/educational institution, 大学/ university, 教育机构/ educational institution, 组织/organization, 学院/institute & 大学/ university, 组织/ organization\\
      \hline
      3. & 对于\textbf{\textcolor{blue}{苹果}}已收购Chomp的报道，Chomp拒加置评，苹果亦尚未就此发表评论。
      
      Regarding the news of \textbf{\textcolor{blue}{Apple}} acquiring Chomp, Chomp refuse to comment, and neither did Apple issue any statement. & 品牌/brand, 公司/company, 上市公司/public company, 科技公司/technology company, 组织/organization & 公司/company, 组织/ organization\\
      \hline
      \midrule
      4. & 对此，德拉吉表示，未与\textbf{\textcolor{blue}{英国央行}}或中国央行在常规操作外进行协作。
      
      Draghi said he did not illegally work with the Bank of England or People's Bank of China. &银行/bank, 政府机构/ government agency , 组织/ organization, 金融机构/financial institution & 银行/bank, 金融机构/financial institution, 政府机构/government agency, 组织/ organization, \textbf{\textcolor{cyan}{金融管理局/monetary authority}}\\
      \hline
      
      
      6. & \textbf{\textcolor{blue}{万里长城}}和太阳金字塔，迄今仍巍然屹立，成为人类文明进步的永恒标志。
      
      \textbf{\textcolor{blue}{The Great Wall}} and the Pyramids are still standing today, becoming a symbol of human civilization. & 地标/landmark, 地点/ location, 旅游景点/ tourist attraction, 文化遗产/cultural heritage, 墙/wall, 位置/location & 地点/location, 旅游景点/tourist attraction, \textbf{\textcolor{cyan}{建筑/architecture}}, \textbf{\textcolor{cyan}{组织/organization}}\\

\bottomrule
       
\end{tabularx}
\caption{\label{tab:res-sent} Test samples of model prediction when training on our distant supervision dataset with general type mapping. Light blue color denotes incorrect predictions.}
\end{center}
\end{table*}
\end{CJK*}
 
\subsection{Training with Distant Supervision Dataset}

We first split the 4,800 crowdsourced examples equally into train, dev and test. Each training batch then comprises equal number of distant supervision data and randomly sampled crowdsourced data from its training set. The development and test set only contain the crowdsourced data. For comparison, we also trained the same model on the Ultra-fine dataset. When training on the Ultra-fine dataset, we followed their original training method, mixing the distant supervision dataset and the crowdsourced dataset to form the training set \cite{choi2018ultra}. The dev set and test set are also only consisting of their crowdsourced data.

\paragraph{BERT}	We use BERT-base-Chinese for our dataset and BERT-base-Cased for the Ultra-fine dataset. We fine-tune BERT on both of the datasets for 5 epochs. We use Adam as optimizer with the learning rate set at 3e-5, $\beta_1=0.9$ and $\beta_2=0.99$. The batch size is 32 and max sequence length is set at 128.

\paragraph{BiLSTM}	We train the whole dataset with bidirectional-LSTM for 15 epochs. The configuration of Adam optimizer is the same as above, with learning rate set at 0.001. We set the batch size at 256 and max sequence length remains the same. 


Both models are tested on our dataset and the Ultra-fine dataset. We also experiment training with and without the general types on our dataset with both models. The evaluation criteria are defined the same as previous work \cite{shimaoka-etal-2016-attentive,choi2018ultra}. Macro-averaged precision, recall, F1-score, and MRR (average mean reciprocal rank) are reported.

\subsection{Training Results and Evaluation}


\begin{table}[h!]
		\begin{center}
			\begin{tabular}{cccc}
				\toprule 
				Level & P & R & F1 \\ 
				\midrule
				General & 79.9 & 74.9 & 77.3\\
				Fine-grained & 28.6 & 22.1 & 24.9\\
				All & 64.1 & 38.2 & 47.9 \\
				\bottomrule
			\end{tabular}
		\end{center}
\caption{\label{tab:fet-break} Breakdown of the prediction results from BERT+General from Table \ref{tab:fet-perf}.}
\end{table}

As shown in Table \ref{tab:fet-perf},  BERT based models show better performance comparing to LSTM based models. Table \ref{tab:fet-break} shows the performance breakdown of different granularity of BERT+General from Table \ref{tab:fet-perf}. 

In most scenarios, we found that the model can predict the general types, but predictions on fine-grained types are more inconsistent. We checked some high-occurrence fine-grained types in our training data and found that the model performs better on them. For example, the type ``writer" has a precision of 0.87 and a recall of 0.72. For low-occurrence types, e.g. ``cultural heritage", the model often fails to predict it.

We inspect some examples of the model predictions on our crowdsourced dataset, as shown in Table \ref{tab:res-sent}. Example 1 shows the case when the model is able to predict correctly, even with a relatively high number of labels (five labels). Example 2 and 3 are situations when entity mentions are labeled more comprehensively, and the model is not able to pick up all the labeled types. The last two examples show situations when the model predicts some types that are not labeled in the ground truth. 

Similar to the Ultra-fine dataset \cite{choi2018ultra}, we find that the type labels of some mentions may be incomplete. This is also similar to a common scenario in recommendation, where only some of the positive examples (the items that users like) are known \cite{heckel2017scalable,pan2008one}. For our data, it is hard to define ``complete" and is almost impossible to construct it for every entity mention. Improving type coverage for each entity mention is an interesting but challenging topic for future work. Nonetheless, our crowdsourced dataset provides high precision on the labeled types, along with a great amount and variety of types for each entity mention. Methods to address the recall issue of incomplete label set should be conducted depending on the use case of this dataset.

Examples in Table \ref{tab:res-sent} show the models are able to learn to predict fine-grained types from our training dataset even with the simplest structures and parameter tunings.

\subsection{Transfer Learning}

Finally, we would like to see whether English fine-grained entity typing data can be used to improve the performance on Chinese data. We experiment transfer learning with Babylon word embedding \cite{smith2017offline} between English and Chinese. We first trained the Ultra-fine dataset on English with the English Babylon word embedding. We then extract the weights of the BiLSTMs and continue training on our Chinese dataset. We experiment training directly on our crowdsourced dataset and also with our distant supervision data. Since we have relatively small number of crowdsourced examples, we split it by a ratio of 8:1:1 for train, dev and test. When training on the distant dataset, we follow our setup in 3.2, splitting the crowdsourced dataset equally to form the train, dev and test set. The results are shown in Table \ref{tab-trans}. All the experiments are conducted with the general type mapping. The result shows improvements under both scenarios. Since most entity typing resources are in English, using transfer learning to improve model performance on low-resource Chinese entity typing tasks is an interesting topic for future work.

\begin{table}[h!]
		\begin{center}
        {
			\begin{tabular}{cp{1cm}cccc}
				\toprule 
				Method & Dataset & MRR & P & R & F1 \\ 
				\midrule

				BiLSTM & crowd & 0.254 & 58.1 & 22.9 & 32.9\\
				BiLSTM + T & crowd & 0.279 & 58.5 & 26.9 & 36.9\\
				\midrule
				\hline
				BiLSTM & distant & 0.200 & 46.6 & 17.5 & 25.5\\
				BiLSTM + T & distant & 0.225 & 57.3 & 22.1 & 31.9 \\
				\bottomrule
			\end{tabular}
        }
		\end{center}
\caption{\label{tab-trans}Experiment results of transfer learning. ``T" indicates transferring the trained BiLSTM weights. ``Dataset" indicates the source of training data.  
Note that the upper half and the lower half are results from different test data and the figures are not comparable between the two halves.
}
\end{table}



\section{Conclusion}
We create a Chinese fine-grained entity typing dataset with each entity mention having an open number of entity types. The dataset contains a large distantly supervised dataset with 1.9M examples, and a smaller crowdsourced dataset containing 4,800 examples with 1,300 unique entity types. In total, our dataset contains 7,100 unique entity types. In addition, a mapping between fine-grained types and general types is established, creating a hierarchical relationship between the large number of types. We test the data on a number of models and show the usability of our dataset.

\section{Acknowledgements}

This paper was supported by the Early Career Scheme (ECS, No. 26206717) from Research Grants Council in Hong Kong and WeChat-HKUST WHAT Lab on Artificial Intelligence Technology.

\section{Bibliographical References}
\label{main:ref}

\bibliographystyle{lrec}
\bibliography{lrec2020W-xample}

\begin{thebibliography}{}

\bibitem[\protect\citename{Abhishek \bgroup et al.\egroup
  }2017]{abhishek-etal-2017-fine}
Abhishek, A., Anand, A., and Awekar, A.
\newblock (2017).
\newblock Fine-grained entity type classification by jointly learning
  representations and label embeddings.
\newblock In {\em Proceedings ACL}, pages 797--807, Valencia, Spain, April.
  Association for Computational Linguistics.

\bibitem[\protect\citename{Chinchor}1998]{chinchor1998overview}
Chinchor, N.
\newblock (1998).
\newblock Overview of muc-7.
\newblock In {\em Proceedings of MUC-7}.

\bibitem[\protect\citename{Choi \bgroup et al.\egroup }2018]{choi2018ultra}
Choi, E., Levy, O., Choi, Y., and Zettlemoyer, L.
\newblock (2018).
\newblock Ultra-fine entity typing.
\newblock In {\em Proceedings of ACL}, pages 87--96.

\bibitem[\protect\citename{Dai \bgroup et al.\egroup
  }2019]{dai-etal-2019-improving}
Dai, H., Du, D., Li, X., and Song, Y.
\newblock (2019).
\newblock Improving fine-grained entity typing with entity linking.
\newblock In {\em Proceedings of EMNLP}, pages 6211--6216, Hong Kong, China,
  November. Association for Computational Linguistics.

\bibitem[\protect\citename{Devlin \bgroup et al.\egroup }2019]{devlin2019bert}
Devlin, J., Chang, M.-W., Lee, K., and Toutanova, K.
\newblock (2019).
\newblock Bert: Pre-training of deep bidirectional transformers for language
  understanding.
\newblock In {\em Proceedings of NAACL}, pages 4171--4186.

\bibitem[\protect\citename{Finkel \bgroup et al.\egroup
  }2005]{finkel2005incorporating}
Finkel, J.~R., Grenager, T., and Manning, C.
\newblock (2005).
\newblock Incorporating non-local information into information extraction
  systems by gibbs sampling.
\newblock In {\em Proceedings of the ACL}, pages 363--370. Association for
  Computational Linguistics.

\bibitem[\protect\citename{Gillick \bgroup et al.\egroup
  }2014]{gillick2014context}
Gillick, D., Lazic, N., Ganchev, K., Kirchner, J., and Huynh, D.
\newblock (2014).
\newblock Context-dependent fine-grained entity type tagging.
\newblock {\em arXiv preprint arXiv:1412.1820}.

\bibitem[\protect\citename{He and Sun}2016]{HeS16}
He, H. and Sun, X.
\newblock (2016).
\newblock F-score driven max margin neural network for named entity recognition
  in chinese social media.
\newblock {\em CoRR}, abs/1611.04234.

\bibitem[\protect\citename{Heckel \bgroup et al.\egroup
  }2017]{heckel2017scalable}
Heckel, R., Vlachos, M., Parnell, T., and D{\"u}nner, C.
\newblock (2017).
\newblock Scalable and interpretable product recommendations via overlapping
  co-clustering.
\newblock In {\em ICDE}, pages 1033--1044. IEEE.

\bibitem[\protect\citename{Ling and Weld}2012]{ling2012fine}
Ling, X. and Weld, D.~S.
\newblock (2012).
\newblock Fine-grained entity recognition.
\newblock In {\em Proceedings of AAAI}.

\bibitem[\protect\citename{Mikolov \bgroup et al.\egroup
  }2018]{mikolov2018advances}
Mikolov, T., Grave, E., Bojanowski, P., Puhrsch, C., and Joulin, A.
\newblock (2018).
\newblock Advances in pre-training distributed word representations.
\newblock In {\em Proceedings of LREC}.

\bibitem[\protect\citename{Mintz \bgroup et al.\egroup
  }2009]{mintz-etal-2009-distant}
Mintz, M., Bills, S., Snow, R., and Jurafsky, D.
\newblock (2009).
\newblock Distant supervision for relation extraction without labeled data.
\newblock In {\em Proceedings of ACL-AFNLP}, pages 1003--1011, Suntec,
  Singapore, August. Association for Computational Linguistics.

\bibitem[\protect\citename{Pan \bgroup et al.\egroup }2008]{pan2008one}
Pan, R., Zhou, Y., Cao, B., Liu, N.~N., Lukose, R., Scholz, M., and Yang, Q.
\newblock (2008).
\newblock One-class collaborative filtering.
\newblock In {\em ICDM}, pages 502--511. IEEE.

\bibitem[\protect\citename{Pennington \bgroup et al.\egroup
  }2014]{pennington2014glove}
Pennington, J., Socher, R., and Manning, C.~D.
\newblock (2014).
\newblock Glove: Global vectors for word representation.
\newblock In {\em Proceedings of EMNLP}, pages 1532--1543.

\bibitem[\protect\citename{Shimaoka \bgroup et al.\egroup
  }2016]{shimaoka-etal-2016-attentive}
Shimaoka, S., Stenetorp, P., Inui, K., and Riedel, S.
\newblock (2016).
\newblock An attentive neural architecture for fine-grained entity type
  classification.
\newblock In {\em Proceedings of AKBC}, pages 69--74, San Diego, CA, June.
  Association for Computational Linguistics.

\bibitem[\protect\citename{Smith \bgroup et al.\egroup }2017]{smith2017offline}
Smith, S.~L., Turban, D.~H., Hamblin, S., and Hammerla, N.~Y.
\newblock (2017).
\newblock Offline bilingual word vectors, orthogonal transformations and the
  inverted softmax.
\newblock {\em arXiv preprint arXiv:1702.03859}.

\bibitem[\protect\citename{Weischedel and Brunstein}2005]{weischedel2005}
Weischedel, R. and Brunstein, A.
\newblock (2005).
\newblock Bbn pronoun coreference and entity type corpus.
\newblock {\em Linguistic Data Consortium, Philadelphia}.

\bibitem[\protect\citename{Yogatama \bgroup et al.\egroup
  }2015]{yogatama2015embedding}
Yogatama, D., Gillick, D., and Lazic, N.
\newblock (2015).
\newblock Embedding methods for fine grained entity type classification.
\newblock In {\em Proceedings of ACL-IJCNLP}, pages 291--296.

\bibitem[\protect\citename{Yu \bgroup et al.\egroup }2018]{DVN/SEYRX5_2018}
Yu, S., Duan, H., and Wu, Y.
\newblock (2018).
\newblock {Corpus of Multi-level Processing for Modern Chinese}.

\end{thebibliography}


\end{document}